\PassOptionsToPackage{table}{xcolor}
\documentclass{selfevolagent}

\usepackage{amsmath}
\usepackage{amssymb}
\usepackage{mathtools}
\usepackage{colortbl}
\usepackage{float}
\usepackage{subcaption}
\usepackage{array}
\usepackage{makecell}
\usepackage{enumitem}
\usepackage{fontawesome5}
\usepackage{listings}

\graphicspath{{./}{figures/}}

\crefname{figure}{Figure}{Figures}
\crefname{table}{Table}{Tables}
\crefname{section}{Section}{Sections}
\crefname{subsection}{Section}{Sections}
\crefname{equation}{Equation}{Equations}
\crefname{appendix}{Appendix}{Appendices}

\newcommand{\method}{SkillGate}
\newcommand{\baserl}{SkillRL\,(outcome only)}
\newcommand{\skillonerow}{Skill1\,(no distill)}
\newcommand{\ourrow}{\rowcolor{gray!15}}
\definecolor{oraclecol}{HTML}{E8F2E4}
\definecolor{misleadcol}{HTML}{FBEBE8}

\title{SkillGate: Training In-Policy Skill Selection in Long-Horizon Agents}

\newcommand{\corrauth}{\text{\raisebox{-0.12ex}{\scalebox{0.78}{\faEnvelope}}}}
\author[1,2\dagger]{Qingyao Li}
\author[2,\corrauth]{~Wenxiang Jiao}
\author[1,2\dagger]{~Shuai Shao}
\author[1,2\dagger]{~Kangning Zhang}
\author[2,\corrauth]{~Yuan Lu}
\author[2]{~Yi Guo}
\author[1,\corrauth]{~Weiwen Liu}
\author[1,\corrauth]{~Weinan Zhang}
\author[1,\corrauth]{~Yong Yu}

\affiliation[1]{Shanghai Jiao Tong University}
\affiliation[2]{Xiaohongshu Inc.}

\contribution[\dagger]{Work done during internship at Xiaohongshu Inc.}
\contribution[\corrauth]{Corresponding authors}

\metadata[\raisebox{-0.18ex}{\scalebox{0.89}{\faEnvelope}}~Contact]{ly890306@sjtu.edu.cn, wenxiangjiaonju@gmail.com, liuww@sjtu.edu.cn}
\metadata[\raisebox{-0.14ex}{\scalebox{0.92}{\faGithub}}~Code]{\url{https://github.com/DeepExperience/SkillGate}}
\metadata[\raisebox{-0.14ex}{\scalebox{0.92}{\faCube}}~Models]{\url{https://huggingface.co/simonlqy/SkillGate-9B}}

\abstract{
Agent frameworks increasingly package procedural knowledge as \emph{skills} ---
instruction files an agent reads on demand --- and public libraries now hold
thousands of them. \emph{Which} skill to read has thus become a decision the policy
itself makes in the middle of an episode, yet no existing signal trains it. We show
that the default remedy --- outcome-rewarded RL over the candidate slate --- cannot
teach it, for a structural reason we identify and name \emph{selector credit
starvation}: under a broadcast, sequence-level advantage, the few tokens that name
the chosen skill carry a vanishing share of the loss, and the credit they inherit
is increasingly wrong-signed as trajectories lengthen --- a correct choice is
punished whenever the execution after it fails --- even though the choice itself is
among the most valuable decisions in the trajectory. Auditing a completed run's own
training artifacts confirms all three properties, each worsening monotonically with
horizon. \method{} removes the failure by construction: it partitions the token
support into two disjoint credit channels, outcome credit reaching only execution
tokens, and a separate action-local advantage reaching exactly the skill-naming
tokens, positive only when a trajectory's single read is the correct one. On five
agentic benchmarks under a 16-candidate slate, \method{} lifts a 9B policy from
$40.8\%$ to $53.2\%$ trial success, well ahead of the identical budget spent on
outcome reward alone, while cutting exposure to misleading candidates by two thirds
and reading \emph{fewer} skills.
}

\begin{document}

\maketitle

\section{Introduction}
\label{sec:intro}

Large Language Models (LLMs) now act in environments rather than only describe them,
using tools, repairing software and controlling computers through multi-turn loops of
reasoning, action and observation \citep{react,toolformer,reflexion,sweagent}, and are
judged on benchmarks that execute their output \citep{agentbench,webarena,osworld,swebench}.
What these agents lack is rarely fluency but \emph{procedural} knowledge, and the
field's answer has been to package it as \emph{skills}: reusable procedural modules,
now commonly exposed through a name and short description before an agent opens their
body on demand \citep{voyager,skillrouter,skillreduce}. Public libraries now hold
thousands of them, far more than a context window admits. At that scale, \emph{when
and which} skill to use becomes the decision that matters, and the agent must make it
from each candidate's name and one-line description alone, in the middle of an
episode, before it can open the file and see what it chose.

Recent work confirms that skill availability alone does not solve this problem.
SRA-Bench \citep{skillra} pairs thousands of tasks with gold skills and finds that base
models often fail to load the right one; Canary Tools \citep{canary} plants decoys that
expose systematic selection errors; and expanding what the agent must choose among can
lower success through skill shadowing or oversized shortlists
\citep{shadowing,slatesize}. Two lines of response have followed. One improves routing
before execution: SkillReducer \citep{skillreduce} and Capability Pages
\citep{cappages} rewrite skill representations, while SkillSight \citep{skillsight},
SkillRet \citep{skillret} and SkillRouter \citep{skillrouter} calibrate, train or
evaluate retrieval. The other optimises skill construction and use together under
task rewards, as in SkillRL \citep{skillrl}, Skill1 \citep{skillone} and SkillRise
\citep{skillrise}. Together these results show that coverage is not enough: which
skill reaches the agent, and how the agent uses it, determine whether the library
helps.

\begin{figure}[t]
\centering
\includegraphics[width=\linewidth]{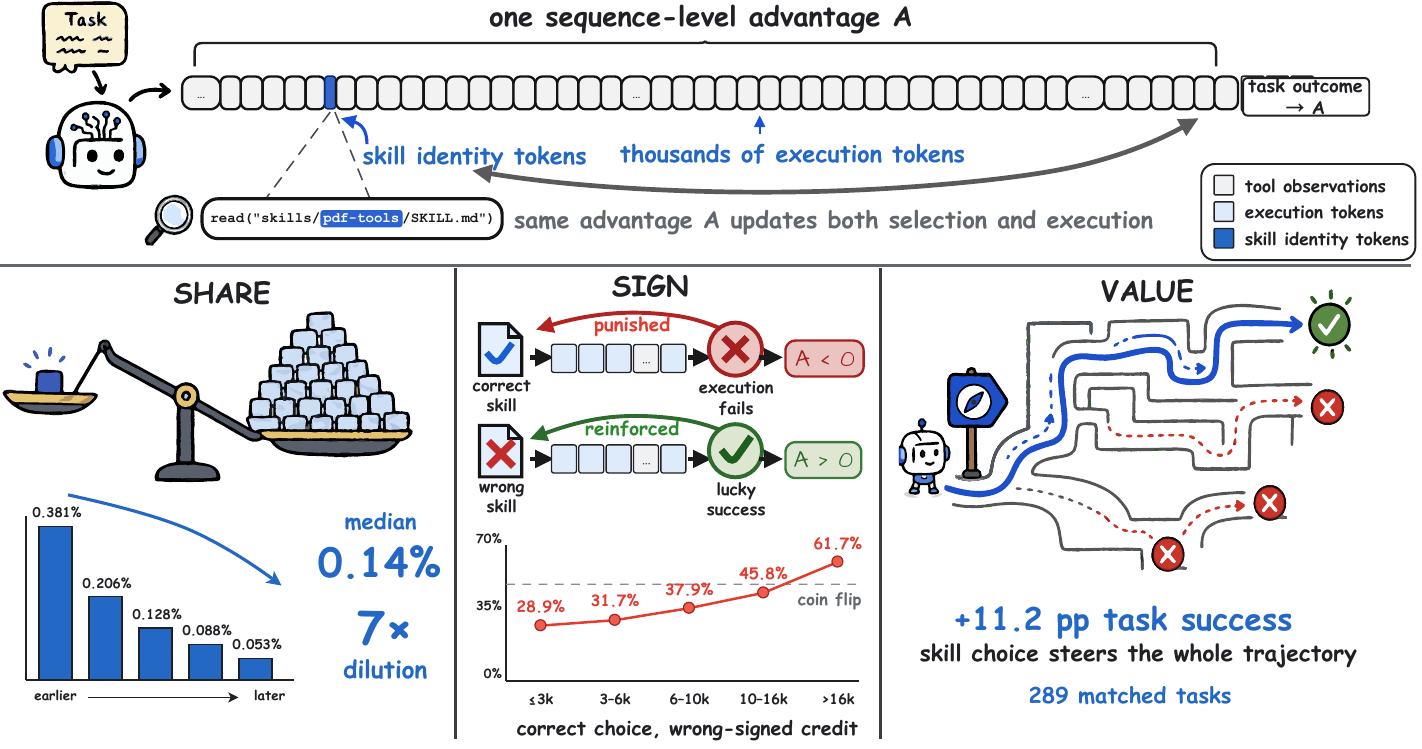}
\caption{Selector credit starvation. One broadcast advantage updates both the few
tokens choosing a skill and the thousands executing the task (\emph{top}); on
12{,}800 training trajectories, the choice's loss share dilutes $7\times$ with
length (\emph{Share}), its credit is increasingly wrong-signed (\emph{Sign}), yet
matched prompt groups show a $+11.2$\,pp success gap for the correct read
(\emph{Value}).}
\label{fig:teaser}
\end{figure}

Despite these advances, \emph{in-policy skill selection} --- the policy's own
mid-episode choice of \emph{which} skill to read --- remains under-addressed. However
the library is curated or ranked beforehand, the read is still issued by the policy
while the task is being solved, and whatever it opens conditions every action after.
Choosing is not easy at that moment either: the candidates are written to look alike,
and a few tokens separate the right name from a plausible wrong one. The obvious way to
train this ability is the objective agentic RL already runs --- present the candidates,
reward the task outcome, and let the policy gradient reach the read
\citep{grpo,ppo,deepseekr1}. We show that in long-horizon tasks this
fails, and fails for a structural reason rather than for want of training, which we
identify as the root cause and name \emph{selector credit starvation}: the few tokens
naming the chosen skill are starved of training signal, and much of what they do
receive pushes the wrong way (\cref{fig:teaser}). Auditing 12{,}800 on-policy trajectories of a run
trained under exactly this recipe makes the failure precise in three respects.
1)~\textbf{Share}: under a broadcast advantage a span's share of the gradient
\emph{equals} its token share, and the tokens naming the skill are a median
$0.14\%$ of a trajectory --- a share that dilutes several-fold as trajectories
lengthen. 2)~\textbf{Sign}: two in five of those tokens receive a \emph{negative}
advantage --- the choice was right, the execution afterwards failed --- and on the
longest trajectories a correct choice is punished more often than rewarded.
3)~\textbf{Value}: within matched prompt groups the correct read is nonetheless
worth $+11.2$ points of task success. A decision this valuable thus receives almost
no signal, wrongly signed much of the time, and all three quantities worsen
monotonically with horizon.

This paper presents a unique perspective on the problem: in-policy skill selection
should be optimised on its own, not left buried in the noise of everything else the
trajectory does. The two decisions are judged by different evidence --- whether the
right file was named is knowable from the slate alone, while whether the work was done
well is knowable only from the outcome --- so pooling them into one advantage lets each
corrupt the other's signal. What is needed is therefore not a better reward, nor a
finer time-resolution of the same reward, but a \emph{separation of credit}: give the
choice its own objective, computed where the choice is visible, and confine the outcome
to the tokens that actually executed.

Building on this perspective, we propose \method{}, which instantiates the partition
as two disjoint credit channels inside one policy-gradient update: (i) a \emph{task
channel} carrying the usual group-normalised outcome advantage, with the entire
skill-read tool call removed so that no task outcome can revise a selection; and
(ii) a \emph{selector channel} applying an action-local advantage to exactly the
tokens naming the skill, positive only when the trajectory issued a single read and
it was the correct one, centred over the prompt group's read actions. The two
supports are disjoint by construction, and both channels carry equal loss weight per
batch with each read action taking an equal share --- so a selection decision's
weight no longer depends on the length of the trajectory containing it. Our
contributions are summarised as follows:

\begin{itemize}[leftmargin=*,itemsep=1pt,topsep=2pt]
\item We identify \emph{selector credit starvation} as the root cause of why
  in-policy skill selection resists outcome-only RL, and --- to our knowledge for the
  first time --- measure it on a real run's own training artifacts: the loss-weight
  share, sign-error rate and within-group signal-to-noise of the selection decision
  all degrade monotonically with trajectory length, while the decision's task value
  does not.
\item We propose \method{}, which partitions one policy's token support into
  disjoint execution and selection credit channels, together with the mechanisms
  that make the partition exact in practice: rollout-time span attribution, a clean
  single-oracle utility with group centring, and length-invariant reweighting.
\item We demonstrate the effectiveness of \method{} on five agentic benchmarks:
  $53.2\%$ trial success against $47.0\%$ for the identical budget spent on outcome
  reward alone, ahead of supervised selection, preference learning, external routers
  and reference models with roughly forty times the parameters. The gain is
  behavioural rather than a capability effect, cutting misleading-skill exposure by
  two thirds while reading \emph{fewer} skills.
\end{itemize}

\section{Problem Setup}
\label{sec:setup}

\paragraph{Skill slate.}
A skill $s$ is a triple $(\mathrm{name}(s), \mathrm{desc}(s), \mathrm{body}(s))$: an
identifier, a one-line description, and a \texttt{SKILL.md} body of typically a few
thousand tokens. Only name and description are shown up front. For each task the
agent sees a \emph{slate} of $K$ candidates, listed in the system prompt and
materialised as files in the sandbox. So that the slate covers the situations a
deployed agent meets, it mixes four kinds of candidate: one \emph{oracle}
$s^\star$, written for the task and verified to solve it; \emph{misleading} hard negatives, topically
adjacent to the task but functionally wrong; and \emph{relevant} and
\emph{irrelevant} bystanders drawn from a public skill library. Nothing is enforced:
the agent is not told which candidate is the oracle, is not required to read
anything, and may read as many as it likes, reading being an ordinary tool call
against a sandbox path. Every number in this paper is measured under this condition,
which we call the \emph{standard mixed slate}; $K$, the per-category counts and each
category's construction are given in \cref{sec:exp:setup}.

\paragraph{Episodes.}
An episode is a multi-turn tool-use loop \citep{react}: the policy $\pi_\theta$ emits
an assistant message with zero or more tool calls, the environment returns
observations, and the loop runs until the agent stops or exhausts a fixed turn and
wall-clock budget. A trajectory $\tau$ concatenates prompt, assistant messages and
observations; only assistant tokens are trained on (mask $m^{\mathrm{base}}_\tau$:
$1$ on generated tokens, $0$ on observations). A terminal verifier returns the task
score $R(\tau) \in [0,1]$ from the benchmark's own tests or grader; it is the only
reward in the system.

\paragraph{Selection and execution.}
Some tool calls open a candidate's body; we call these \emph{read actions}, writing
$A(\tau)$ for the ordered list of them in $\tau$. A read is attributed only from
assistant-generated text --- never from an observation, which could echo a path ---
and only when it opens a file under a skill directory, whether through the
\texttt{read} tool or a shell command; each attributed read therefore carries a
skill identity and a slate category. Within a read $a$ we record, against the
training tokenizer, the \emph{call span} $C(a)$ (the whole tool call) and the
nested \emph{identity span} $I(a) \subseteq C(a)$ (the skill name inside the path);
a trajectory whose spans fail to align is failed closed rather than silently
mis-credited. On these
spans we impose the division this paper turns on --- ours, not a natural
segmentation of the rollout: \emph{selection} is the identity spans, a handful of
tokens settled early whose correctness is knowable from the slate alone;
\emph{execution} is every other assistant token, whose quality is knowable only from
the outcome.

\section{\method{}}
\label{sec:method}

\begin{figure}[t]
\centering
\includegraphics[width=\linewidth]{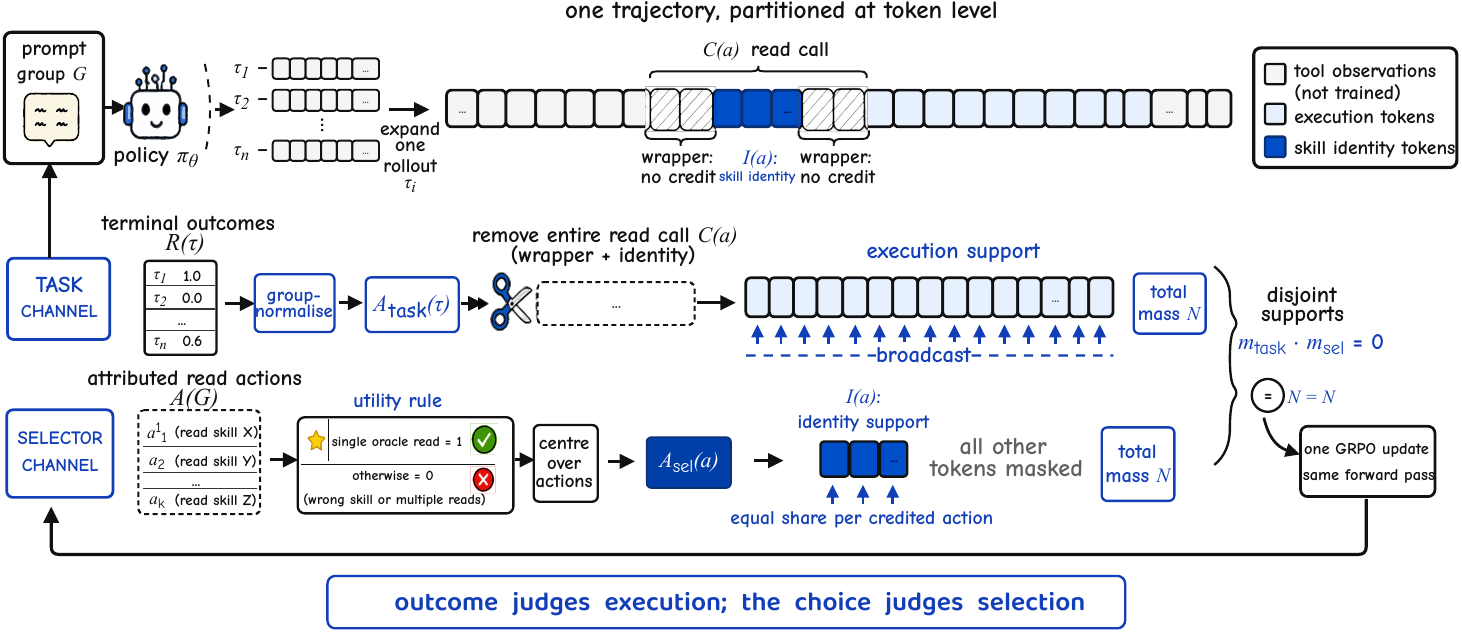}
\caption{\method{} overview. The task channel group-normalises terminal outcomes
into $A^{\mathrm{task}}$ and broadcasts it over execution tokens with the entire
read call cut out; the selector channel scores read actions with the clean
single-oracle utility and places $A^{\mathrm{sel}}$ on the identity tokens alone.
The supports are disjoint and each channel's token weights sum to $N$ before
$\lambda$ is applied; both enter one GRPO update.}
\label{fig:method}
\end{figure}

\method{} trains a single policy to do two things well at once: \emph{select} the
right skill, and \emph{execute} the task with what it read. It does so through two
credit channels that never touch the same token --- execution tokens receive the
sequence-level task advantage, while the tokens naming the chosen skill receive an
action-local advantage that depends on the choice and on nothing else
(\cref{fig:method}). Rollout, reward and optimiser are untouched: everything in
this section is a statement about \emph{where} an advantage is allowed to land.

\subsection{Token-Level Credit Assignment}
\label{sec:method:credit}

\method{} splits the trained tokens of a trajectory along the read-action spans of
\cref{sec:setup}:
\begin{itemize}[leftmargin=*,itemsep=1pt,topsep=2pt]
\item \textbf{Selection tokens}, $\bigcup_{a \in A(\tau)} I(a)$: the tokens that
  write a skill's name. They are the entire surface of the selection decision ---
  change them, and a different file is read.
\item \textbf{Execution tokens}, every assistant token outside
  $\bigcup_{a \in A(\tau)} C(a)$: the reasoning, the other tool calls, the work.
\end{itemize}
What remains of a call span once its identity span is removed --- the tool-call
wrapper --- belongs to the selection decision but not to its identity, and is
trained by neither channel; skill bodies arrive as observations and are never
trained. Each kind of token then earns its own credit. Training proceeds in prompt
groups --- for each task the policy produces a group $G$ of $n$ independent
rollouts --- and:
\begin{itemize}[leftmargin=*,itemsep=2pt,topsep=2pt]
\item \textbf{Outcome credit} judges execution by whether the task succeeded,
  relative to the group: the usual group-normalised GRPO advantage \citep{grpo},
  $A^{\mathrm{task}}(\tau) = \big(R(\tau)-\mu_G\big)/\big(\sigma_G+\epsilon\big)$,
  constant along the trajectory, where $\mu_G$ and $\sigma_G$ are the mean and
  standard deviation of the task scores $\{R(\tau)\}_{\tau \in G}$ and $\epsilon$
  is a small constant.
\item \textbf{Selection credit} judges each read action by the choice it made and
  nothing else. Writing $A(G)$ for the read actions of the whole group, an action's
  utility asks one question and its advantage is the group-centred answer:
  \begin{equation}
  u(a) =
  \begin{cases}
  1, & |A(\tau_a)| = 1 \ \text{and $a$ reads } s^\star,\\
  0, & \text{otherwise,}
  \end{cases}
  \quad
  A^{\mathrm{sel}}(a) = u(a) - \tfrac{1}{|A(G)|}\,\textstyle\sum_{a' \in A(G)} u(a'),
  \label{eq:sel}
  \end{equation}
  where $\tau_a$ is the trajectory containing $a$, and no standard-deviation
  normalisation is applied --- action counts are small, and dividing by their
  spread would amplify noise more than calibrate scale. Reading the oracle and then
  three more candidates scores zero, as does reading it twice: the single-read
  requirement is what stops the policy from buying credit by reading everything.
\end{itemize}

Three properties of \cref{eq:sel} bound what the selector term can do. The baseline
is over \emph{actions}, not trajectories, so a promiscuous sibling that reads four
candidates lowers it for everyone. The action-weighted sum of $A^{\mathrm{sel}}$
over a group is exactly zero, so the term exerts no standing pressure for or
against reading in general --- only over \emph{which} name is written. And it is
self-limiting: if a group contains no clean oracle read, or only clean oracle
reads, every utility ties, $A^{\mathrm{sel}} \equiv 0$, and the channel falls silent. Signal
appears only where the group disagrees.

\subsection{Credit Masking and Normalisation}
\label{sec:method:masks}

The two channels are kept apart by construction. The task channel deletes every
read call \emph{whole} from the assistant mask --- wrapper, function name and path,
not merely the skill name --- so that no task outcome can revise a selection; what
survives is exactly the execution tokens. The selector channel trains only the
identity spans of credited actions (those with $A^{\mathrm{sel}}(a) \neq 0$), every
token of a span carrying its action's advantage. Because identity spans lie inside
the deleted call spans, the two supports are disjoint; the implementation asserts
this at three points --- mask construction, context-parallel sharding, and the loss
itself --- and any violation aborts training rather than degrading it silently. An
off-slate read is deleted from the task mask like any other read but earns no
selector credit, so it cannot be laundered into a positive update by a lucky
outcome.

Token weights then set how loudly each channel speaks, and the two need rescaling
for opposite reasons. Let $m^{\mathrm{base}}_t$ be the original assistant-token
loss mask and write $N=\sum_t m^{\mathrm{base}}_t$ for its total weight in a batch.
After deleting all read calls, let $N_{\mathrm{task}}=\sum_t
m^{\mathrm{task}}_t$, and let $M$ be the number of credited read actions. We set
\begin{equation}
w^{\mathrm{task}}_t = \frac{N}{N_{\mathrm{task}}}m^{\mathrm{task}}_t,
\qquad
w^{\mathrm{sel}}_t = \frac{N}{M\,\lvert I(a)\rvert}
\quad \text{for }t\in I(a),
\label{eq:mass}
\end{equation}
with zero selector weight elsewhere. When $M>0$, hence $\sum_t
w^{\mathrm{task}}_t=N$ and $\sum_t w^{\mathrm{sel}}_t=N$: this is the
``$N=N$'' in \cref{fig:method}; when $M=0$, the selector channel is silent.
Deleting read calls therefore does not quietly lower the task channel's effective
learning rate, while the handful of selector tokens no longer reproduces the
starvation of \cref{fig:teaser}. The equality concerns total token-weight mass,
not equality of the realised loss values or gradient norms; the intended relative
coefficient is still set by $\lambda$ in \cref{eq:objective}. It also gives every
credited action the same total weight $N/M$, independent of trajectory length and
of how many tokenizer pieces its skill name occupies. The implementation verifies
both sums numerically every batch.

\subsection{Training Objective}
\label{sec:method:objective}

Let $r_t(\theta) = \pi_\theta(y_t \mid y_{<t}) / \pi_{\mathrm{rollout}}(y_t \mid y_{<t})$
be the importance ratio of token $y_t$ against the rollout policy, and
$\ell^{\mathrm{clip}}(r, A) = -\min\big(rA,\ \mathrm{clip}(r, 1-\varepsilon_{\mathrm{lo}},
1+\varepsilon_{\mathrm{hi}})A\big)$ the clipped GRPO surrogate \citep{grpo} with
clip thresholds $\varepsilon_{\mathrm{lo}}, \varepsilon_{\mathrm{hi}}$. \method{}
optimises
\begin{equation}
\mathcal{L}(\theta)
= \underbrace{\sum_{t} w^{\mathrm{task}}_t\, \ell^{\mathrm{clip}}\!\big(r_t(\theta), A^{\mathrm{task}}(\tau_t)\big)}_{\text{execution}}
\;+\; \lambda \underbrace{\sum_{t} w^{\mathrm{sel}}_t\, \ell^{\mathrm{clip}}\!\big(r_t(\theta), A^{\mathrm{sel}}(a_t)\big)}_{\text{selection}}
\;+\; \beta\, \mathcal{L}_{\mathrm{KL}},
\label{eq:objective}
\end{equation}
where $\tau_t$ and $a_t$ are the trajectory and the read action containing token
$t$, $\lambda$ is the selector coefficient, and $\mathcal{L}_{\mathrm{KL}}$ is a KL
penalty of weight $\beta$ against a frozen reference (values in
\cref{sec:exp:setup}). Both sums run over the same forward pass and both terms are
on-policy: the selector term is not behaviour cloning, not cross-entropy and not a
reward bonus, and it never sees a token the policy did not itself generate.

The credit semantics are then exhaustive over the four cases that matter. A
trajectory that reads only the oracle raises that name whether the task succeeded or
failed; one that reads a misleading candidate lowers that name in either case; and
the execution tokens are rewarded or penalised on outcome regardless of what was
read. A negative value in \method{}'s selector term therefore means \emph{the wrong
skill} or \emph{more than one skill} --- never \emph{the oracle's content did not
work}.

\section{Experiments}
\label{sec:exp}

\subsection{Experimental Setup}
\label{sec:exp:setup}

\paragraph{Datasets.}
We evaluate on five agentic benchmarks --- Claw-Eval, SkillsBench, SETA, SWE and
Terminal-Bench~2.0 --- under the standard mixed slate of \cref{sec:setup} with
$K=16$: one \emph{oracle} written for the task, five \emph{misleading} hard
negatives synthesised to be topically adjacent but functionally wrong, and five
\emph{relevant} plus five \emph{irrelevant} skills sampled from a public library
of 2{,}045 community skills. Slate order is randomised per task, and each trial
runs under a budget of 30 turns and 850 seconds. Training uses 491 tasks
containing \emph{no} Claw-Eval task, and the training and evaluation oracles are
disjoint as skill identities, so no result can be explained by memorising which
name goes with which task. Task outcome is reported on a \textbf{385-trial}
protocol: four repeats for each of 56 non-Claw tasks and one trial for each of
161 Claw-Eval tasks. \Cref{app:unified} describes its construction.
Read behaviour in \cref{tab:main} is measured on those same 385 trials, so
outcome and behaviour share a denominator; the finer behaviour breakdowns, which
need per-trial attribution across repeats, use the \textbf{280-trial} subset (70
tasks with four repeats each) and are labelled as such.

\paragraph{Baselines.}
We compare \method{} against four categories of methods:
\begin{itemize}[leftmargin=*,itemsep=1pt,topsep=2pt]
\item \textbf{Untrained backbones}: \textbf{Qwen3.5-9B} and \textbf{Qwen3.5-27B}
  bracket what the backbone does without adaptation; the 27B row is a capability
  reference at $3\times$ the parameters.
\item \textbf{Frontier references}: DeepSeek-V4-Pro, DeepSeek-V4-Flash, GLM-5,
  Kimi-K2.6, Qwen3.5-397B-A17B and DeepSeek-V3.2, served through their
  providers' APIs with native function calling. They put the trained rows on an
  absolute scale and are not controlled comparisons; \cref{app:frontier} states
  the interface controls used for these reference models.
\item \textbf{Supervised selection}: \textbf{Selection BC} teacher-forces the
  first turn to \texttt{read(oracle)} on all 491 training tasks;
  \textbf{SelSkill-DPO} adapts the invoke-versus-skip preference method of
  \citet{selskill} to our 16-way slate by preferring \texttt{read(oracle)} over
  each hard negative (2{,}455 pairs) under the DPO objective \citep{dpo} --- an
  adaptation, not a reproduction.
\item \textbf{RL with outcome reward}: \textbf{Skill-free RL} trains without
  skills; \textbf{\baserl{}} trains on the same slate with task reward alone ---
  an outcome-only skill-augmented RL recipe inspired by \citet{skillrl}, rather
  than a reproduction of their full skill-evolution system, and the controlled comparison for \method{}
  (same initialisation, data, steps and hyperparameters, differing only in what
  the gradient reaches); \textbf{\skillonerow{}} adapts the joint selection-and-use
  component of \citet{skillone}, omitting its distillation stage so that it too
  shares our initialisation and budget;
  \textbf{Task-mask only} is \method{} with the selector coefficient set to zero.
\end{itemize}

\paragraph{Evaluation Metrics.}
We assess performance along two axes:
\begin{itemize}[leftmargin=*,itemsep=1pt,topsep=2pt]
\item \textbf{Trial success (\%)}: the fraction of trials whose terminal
  verifier passes --- the primary metric, per benchmark and pooled.
\item \textbf{Read behaviour}: \emph{oracle} and \emph{misleading exposure}
  (fraction of trials reading at least one skill of that category; a trial can
  count in both), $P(\text{oracle}\mid\text{read})$, \emph{reads/trial}
  (distinct skill names), and \emph{clean single-oracle} --- exactly one
  attributed read and it is the oracle, the quantity \method{}'s utility is
  defined on.
\end{itemize}

\paragraph{Implementation Details.}
All trained rows start from the same Qwen3.5-9B SFT checkpoint, which is also
the RL initialisation. RL runs 100 steps of on-policy GRPO \citep{grpo} on the
491 training tasks with 8 rollouts per prompt, a global batch of 128
trajectories, learning rate $10^{-6}$ and KL coefficient $3\times10^{-5}$;
\method{} adds the selector term of \cref{sec:method} at $\lambda=0.20$. Each
configuration is a single run (100 steps on 16 H800s), so uncertainty is
quantified by task-level bootstrap rather than seed replication
(\cref{app:bootstrap}) and small differences are read as directional.

\subsection{Main Results}
\label{sec:exp:main}

\begin{table}[t]
\centering
\caption{Trial success (\%) on the standard mixed slate, and read behaviour.
\emph{Oracle}/\emph{misleading}: fraction of trials reading at least one such
skill (a trial can read both). Best 9B-scale per column in \textbf{bold}.}
\label{tab:main}
\footnotesize
\setlength{\tabcolsep}{5pt}
\begin{tabular}{lccccc@{\hskip 8pt}c@{\hskip 8pt}cc}
\toprule
& \multicolumn{6}{c}{Trial success (\%)} & \multicolumn{2}{c}{Read behaviour (\%)} \\
\cmidrule(lr){2-7}\cmidrule(l){8-9}
Method & Claw-Eval & SkillsBench & SETA & SWE & TB2 & Overall
  & \cellcolor{oraclecol}Oracle $\uparrow$ & \cellcolor{misleadcol}Mislead.\ $\downarrow$ \\
\midrule
\multicolumn{9}{l}{\emph{Frontier models (capability ceiling, not controlled)}}\\
DeepSeek-V4-Flash & 70.2 & 15.6 & 66.7 & 55.0 & 46.9 & 61.0 & \cellcolor{oraclecol}40.8 & \cellcolor{misleadcol}23.9 \\
GLM-5 & 75.8 & 15.6 & 59.2 & 45.0 & 56.2 & 60.8 & \cellcolor{oraclecol}39.0 & \cellcolor{misleadcol}26.2 \\
DeepSeek-V4-Pro & 70.8 & 12.5 & 60.8 & 55.0 & 62.5 & 60.5 & \cellcolor{oraclecol}38.2 & \cellcolor{misleadcol}20.8 \\
Kimi-K2.6 & 71.4 & 12.5 & 56.7 & 55.0 & 71.9 & 60.3 & \cellcolor{oraclecol}45.7 & \cellcolor{misleadcol}21.0 \\
Qwen3.5-397B-A17B & 60.9 & 15.6 & 54.2 & 45.0 & 40.6 & 51.7 & \cellcolor{oraclecol}16.1 & \cellcolor{misleadcol}11.2 \\
DeepSeek-V3.2 & 62.1 & 12.5 & 42.5 & 32.5 & 46.9 & 47.5 & \cellcolor{oraclecol}29.4 & \cellcolor{misleadcol}28.3 \\
\midrule
\multicolumn{9}{l}{\emph{No adaptation}}\\
Qwen3.5-9B & 44.7 & \phantom{0}0.0 & 24.2 & 15.0 & \phantom{0}9.4 & 28.6 & \cellcolor{oraclecol}\phantom{0}5.7 & \cellcolor{misleadcol}\phantom{0}8.9 \\
Qwen3.5-27B & 54.0 & \phantom{0}9.4 & 42.5 & 40.0 & 34.4 & 43.6 & \cellcolor{oraclecol}\phantom{0}5.7 & \cellcolor{misleadcol}\phantom{0}8.9 \\
\midrule
\multicolumn{9}{l}{\emph{Supervised selection, from SFT}}\\
SFT (RL init) & 50.9 & \phantom{0}6.2 & 40.0 & 45.0 & 21.9 & 40.8 & \cellcolor{oraclecol}37.9 & \cellcolor{misleadcol}61.8 \\
Selection BC & 52.2 & \textbf{15.6} & 43.3 & 50.0 & 34.4 & 44.7 & \cellcolor{oraclecol}71.4 & \cellcolor{misleadcol}31.4 \\
SelSkill-DPO & 52.8 & \phantom{0}0.0 & 47.5 & 60.0 & \textbf{37.5} & 46.2 & \cellcolor{oraclecol}66.1 & \cellcolor{misleadcol}51.8 \\
\midrule
\multicolumn{9}{l}{\emph{RL from SFT, 100 steps, identical budget}}\\
Skill-free RL & 55.9 & \phantom{0}9.4 & 47.5 & 42.5 & 31.2 & 46.0 & \cellcolor{oraclecol}35.4 & \cellcolor{misleadcol}55.0 \\
\baserl{} & 57.1 & \phantom{0}3.1 & 50.0 & 45.0 & 31.2 & 47.0 & \cellcolor{oraclecol}54.3 & \cellcolor{misleadcol}69.6 \\
\skillonerow{} & 57.1 & \phantom{0}9.4 & 38.3 & 52.5 & 31.2 & 44.7 & \cellcolor{oraclecol}53.5 & \cellcolor{misleadcol}45.5 \\
Task-mask only & 54.7 & \phantom{0}9.4 & 48.3 & 45.0 & 31.2 & 46.0 & \cellcolor{oraclecol}48.9 & \cellcolor{misleadcol}73.6 \\
\ourrow \textbf{\method{}} & \textbf{60.2} & \textbf{15.6} & \textbf{54.2} & \textbf{65.0} & \textbf{37.5} & \textbf{53.2} & \cellcolor{oraclecol}\textbf{83.9} & \cellcolor{misleadcol}\textbf{21.8} \\
\bottomrule
\end{tabular}
\end{table}

\paragraph{\method{} delivers the strongest task performance at the 9B scale.}
As shown in \cref{tab:main}, \method{} outperforms the shared SFT initialisation,
the controlled outcome-only RL run, and the supervised, preference-based and
Skill1 alternatives. It is best or tied best among the 9B-scale methods on every
benchmark. The comparison with \baserl{} is especially informative because the
two runs share their initialisation, data, update budget and hyperparameters;
their difference isolates the value of token-local selection credit. \method{}
also leads the 9B rows on Claw-Eval despite seeing neither Claw tasks nor the
evaluation oracle identities during training, supporting transfer rather than a
memorised task--skill mapping.

\paragraph{Reliable skill access is consequential, not merely a behaviour metric.}
The oracle-only intervention in \cref{tab:selector} improves the frozen SFT
executor by roughly eleven points, establishing that access to the correct skill
materially affects task success. Yet outcome-only RL raises both oracle and
misleading exposure: it learns to read more, not to discriminate better. Simply
masking read calls from the task loss does not fix the problem either. \method{}
instead produces the strongest joint shift toward oracle reads and away from
misleading reads, and converts that behavioural improvement into task success.
Selection BC improves the choice but not the downstream result to the same degree,
showing that reliable agents require selector learning and execution learning
together.

\paragraph{Model scale alone does not solve in-policy skill selection.}
The frontier rows are scale references rather than controlled comparisons, but
they reveal a clear separation between general capability and skill selection.
\method{} surpasses Qwen3.5-397B-A17B and DeepSeek-V3.2 while remaining below the
four strongest frontier systems on task success. More importantly, none of the
frontier models reads the oracle on even half of the trials. Their general
capability can compensate for many missed skills, but it does not produce a
reliable selector. The contrast with \method{} shows that choosing the right skill
is a distinct capability that must be trained directly rather than expected to
emerge from scale.

\subsection{Which part of the credit design matters}
\label{sec:exp:ablation}

\begin{table}[t]
\centering
\caption{Credit-design ablation, 280-trial protocol: five placements of the
selector signal from the same initialisation and steps. \emph{Clean
single-oracle}: exactly one attributed read and it is the oracle. Best results
in \textbf{bold}.}
\label{tab:ablation}
\footnotesize
\setlength{\tabcolsep}{5pt}
\begin{tabular}{llccccc}
\toprule
Design & Where the credit lands & \makecell{Trial\\succ.\ (\%)} & \makecell{Clean single-\\oracle (\%)} & \makecell{Oracle\\(\%)} & \makecell{Mislead.\\(\%)} & \makecell{Reads/\\trial} \\
\midrule
\baserl{} & nowhere (anchor) & 42.1 & 21.4 & 54.3 & 69.6 & 1.88 \\
Group-level regret & the prompt group & 41.8 & 15.7 & 33.6 & 47.5 & 1.23 \\
Trajectory bonus & the whole trajectory & 41.8 & 33.9 & 47.9 & 55.4 & 1.33 \\
Action credit & first oracle read & 45.0 & 64.6 & 80.0 & 32.5 & 1.26 \\
\ourrow \textbf{\method{}} & the \emph{only} read, if oracle & \textbf{50.0} & \textbf{75.4} & \textbf{83.9} & \textbf{21.8} & \textbf{1.11} \\
\bottomrule
\end{tabular}
\end{table}

\begin{figure}[t]
\centering
\includegraphics[width=0.86\linewidth]{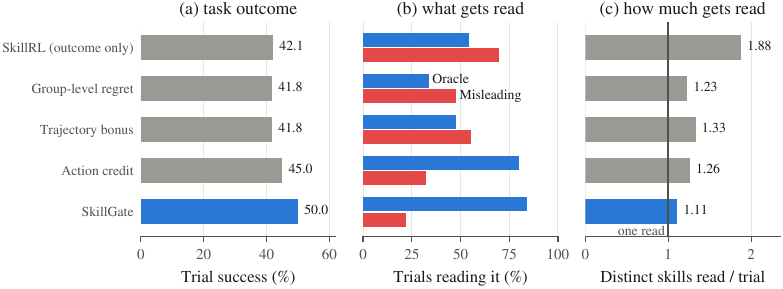}
\caption{The five credit designs of \cref{tab:ablation}: task outcome (a), what
gets read (b), and how much gets read (c). Only credit that reaches the identity
tokens changes the choice, and the single-read rule converts it into success.}
\label{fig:ablation}
\end{figure}

This ablation asks which part of the credit design does the work. All five rows
of \cref{tab:ablation} train from the same initialisation for the same 100 steps
and differ only in where the selector signal lands: nowhere (\baserl{}, the
anchor), on the whole prompt group, on the whole trajectory, on the first oracle
read, or --- \method{} --- on the trajectory's only read when it is the oracle.
\cref{fig:ablation} plots the same five designs; both use the 280-trial
protocol.

Each coarser placement fails in a way that identifies what the next must fix.
Group-level regret never touches the tokens that name the skill, and its oracle
exposure falls below even the anchor's: a group-constant shift cancels within
the group and cannot say which member chose well. The trajectory bonus pushes
the policy to read, not to read correctly --- oracle and misleading exposure
rise together. Action credit finally lands the signal on the identity tokens and
moves both behaviour columns sharply, but it never penalises reading
\emph{more}: a trajectory that reads the oracle and then three other skills
keeps its full credit, so clean single-oracle behaviour stalls. Requiring the
read to be the trajectory's only one closes that gap and converts it into task
success. \cref{fig:ablation} makes the mechanism visible: \method{} not only
reads the oracle more often but reads misleading candidates less (b) and reads
fewer skills overall (c) --- the gain comes from choosing better, not from
reading more.

\subsection{Would a separate selector be enough?}
\label{sec:exp:selector}

\begin{table}[t]
\centering
\caption{External selection, 280-trial protocol. Routers advertise one chosen
skill to the frozen SFT executor; oracle-only injection is a ceiling that
requires knowing the answer. $^\dagger$Top-1 is over 70 routing decisions for
routers but \method{}'s first attributed read over 280 trials --- not directly
comparable.}
\label{tab:selector}
\footnotesize
\setlength{\tabcolsep}{4pt}
\begin{tabular}{lccccc}
\toprule
Setting & \makecell{Trial\\succ.\ (\%)} & \makecell{Top-1\\oracle$^\dagger$ (\%)} & \makecell{Oracle\\(\%)} & \makecell{Mislead.\\(\%)} & \makecell{Reads/\\trial} \\
\midrule
SFT, standard mixed slate      & 37.1 & --- & 37.9 & 61.8 & 1.34 \\
SFT $+$ ``read only one'' prompt & 35.0 & --- & 41.4 & 59.3 & 1.35 \\
\midrule
SFT-9B router $\to$ SFT        & 40.7 & 60.0 & 57.5 & 36.8 & 0.94 \\
Qwen3.5-27B router $\to$ SFT   & 36.8 & 68.6 & 68.2 & 24.3 & 0.95 \\
Qwen3 reranker top-1 $\to$ SFT & 31.8 & 27.1 & 24.6 & 60.4 & 0.89 \\
\midrule
\ourrow \textbf{\method{}}, standard mixed slate & \textbf{50.0} & 78.9 & 83.9 & 21.8 & 1.11 \\
\midrule
\multicolumn{6}{l}{\emph{Ceiling: only the correct skill is advertised}}\\
oracle-only $\to$ SFT          & 48.2 & 100.0 & 93.6 & \phantom{0}0.0 & 0.94 \\
oracle-only $\to$ \method{}    & 52.9 & 100.0 & 93.6 & \phantom{0}0.0 & 0.94 \\
\bottomrule
\end{tabular}
\end{table}

This experiment asks whether selection is better solved outside the policy.
\cref{tab:selector} varies who chooses while freezing the executor: the SFT
model under the standard slate, with and without a ``read only one'' system
prompt; three external selectors --- an SFT-9B router, a Qwen3.5-27B router and
a retrieval reranker --- each advertising its single chosen skill to the frozen
SFT executor; \method{}, choosing by itself from the full slate; and an
oracle-only injection that advertises the correct skill alone, a ceiling that
requires knowing the answer.

Three conclusions follow from \cref{tab:selector}. First, prompting does not
fix selection: instructing the model to read exactly one skill changes neither
its reading nor its success, so the behaviour is not an instruction-following
failure. Second, a better router is not a better agent: the 27B router routes
more accurately than the SFT router yet yields \emph{lower} downstream success
--- routing accuracy, whether the executor reads what it is handed, and how well
it executes afterwards are three different quantities --- and the reranker,
scoring candidates independently rather than comparing them, is precisely what
a slate of near-duplicates defeats. Third, the ceiling is closer than the
routers get: \method{}, choosing from all 16 candidates itself, surpasses even
the oracle-only ceiling of the frozen executor, and applying the same injection
on top of \method{} still helps --- so selection training did not damage the
policy's ability to use a skill once it has one, and only a few points of
selection headroom remain.

\subsection{What selection purity costs}
\label{sec:exp:cost}

\begin{figure}[t]
\centering
\includegraphics[width=0.86\linewidth]{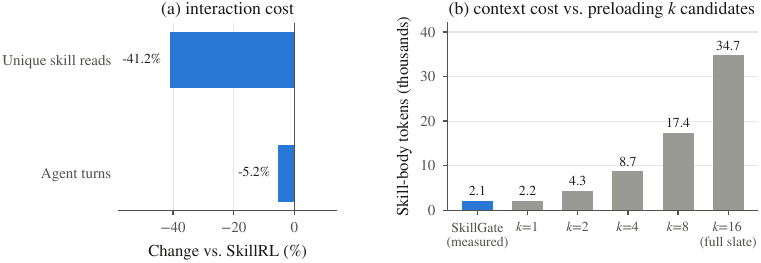}
\caption{Inference cost, 280-trial protocol. (a) \method{} vs.\ \baserl{}:
reads, turns and token usage. (b) Skill-body tokens loaded: measured on-demand
reading vs.\ preloading $k$ candidates' \texttt{SKILL.md} bodies.}
\label{fig:cost}
\end{figure}

This section measures what selection purity costs at inference time.
\cref{fig:cost}(a) compares \method{} with the outcome-only baseline on the same
280 trials
along reads, turns and token usage; \cref{fig:cost}(b) compares \method{}'s
measured on-demand skill-body tokens with hypothetically preloading $k$
candidates' bodies into context.

As \cref{fig:cost}(a) shows, \method{} is cheaper than its outcome-only
counterpart on every count --- fewer distinct reads, fewer turns, fewer
cumulative input tokens --- with only output tokens rising, consistent with
reasoning before a single read rather than trial-and-error after many. On
context cost (\cref{fig:cost}b), \method{}'s on-demand reading loads about as
many body tokens as preloading exactly one skill would, without an external
component to say which one; preloading grows linearly in $k$ and reaches
$16\times$ the on-demand cost at the full slate.

\subsection{Why outcome-only credit cannot teach the choice}
\label{sec:exp:mechanism}

\begin{figure}[t]
\centering
\includegraphics[width=\linewidth]{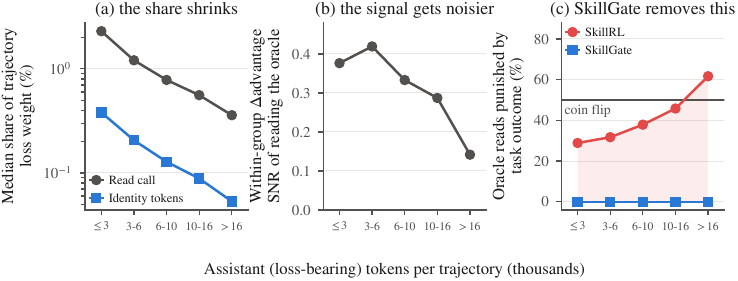}
\caption{Offline audit of \baserl{}'s 12{,}800 training trajectories, by
trajectory-length bin. (a) Median loss-weight share of the read call and of the
skill-naming tokens. (b) Within-group signal-to-noise of the read-vs-not
advantage gap. (c) Share of oracle reads inheriting a negative advantage;
\method{}'s zero is by construction.}
\label{fig:starvation}
\end{figure}

This section verifies the diagnosis that motivated \method{}: that under
outcome-only credit, the selection decision is starved of signal. The audit is
offline and runs no new training: we re-read \baserl{}'s own 100 steps and
12{,}800 on-policy trajectories, recompute the GRPO advantage under the training
normalisation, locate the skill-identity tokens with the training tokenizer, and
stratify by trajectory length. Results are shown in \cref{fig:starvation}.

\paragraph{The decision carries almost none of the gradient, and less as
trajectories grow.}
Because the advantage is constant within a trajectory and the loss averages over
each trajectory's mask, a span's share of the gradient equals its token share.
As \cref{fig:starvation}(a) shows, the tokens naming the skill carry a median
$0.14\%$ of their trajectory's loss weight, and the share dilutes roughly
$7\times$ from the shortest length bin to the longest. This is a statement about
loss weight, not gradient norm.

\paragraph{What little signal arrives is increasingly the wrong sign.}
Nearly two in five oracle-reading trajectories inherit a \emph{negative}
advantage because the executor later did worse than its group-mates: the policy
is told not to make a choice it made correctly. Stratified by length, the rate
climbs monotonically and crosses the coin-flip line on the longest trajectories,
while the within-group signal-to-noise of the same comparison collapses
(\cref{fig:starvation}b,c).

\paragraph{Meanwhile the decision is worth having.}
Within prompt groups that contain both kinds of rollout --- same task, same
step, same policy snapshot --- reading the oracle is worth $+11.2$\,pp of
success. A decision this valuable thus receives a vanishing, increasingly
wrong-signed signal precisely where trajectories are longest. The comparison is
behaviour-conditioned rather than randomised; its causal counterpart, the
oracle-only injection of \cref{tab:selector}, is worth a nearly identical
amount to the frozen executor.

\paragraph{\method{} removes exactly this failure, by construction.}
With the read call excluded from the task loss, the share of oracle identity
tokens receiving a negative task gradient is zero at every length
(\cref{fig:starvation}c). The selector channel still delivers negatives, but
only for the intended reason --- the wrong skill, or more than one read ---
never because the oracle's content did not work.

\section{Related Work}
\label{sec:related}

\paragraph{Selecting a skill from a large library.}
Agent frameworks expose skills by \emph{progressive disclosure} --- the agent sees a
name and a one-line description and decides what to open --- and recent work shows the
difficulty concentrates there. Audits of public libraries find a large share of skills
carry no usable routing description at all \citep{skillreduce}, and dedicated
benchmarks now measure the resulting confusion at scale
\citep{skillret,skillra,skillresolvebench}. Controlled studies isolate the failure:
enlarging a library degrades agents through \emph{skill shadowing} \citep{shadowing},
planted decoy tools expose systematic selection errors \citep{canary}, attention
analyses find that models often attend to the correct tool definition yet select a
different tool at readout \citep{lookpick}, and both shortlist size and
progressive-disclosure depth have become objects of study
\citep{slatesize,progdisc}; even whether a tool is needed at all has its own benchmark
\citep{when2tool}. One response improves routing before execution by rewriting skill
representations \citep{cappages,skillreduce}, calibrating retrieval scores
\citep{skillsight}, or retrieving and reranking candidates
\citep{rag,toolllm,gorilla,skillrouter}. Benchmarks separately diagnose skill
triggering and use \citep{skilluse} or large-scale tool retrieval and navigation
\citep{mcpco}. Methods that explicitly optimise selection often give it a separate
training structure: AutoTool uses a KL-regularised Plackett--Luce ranking objective
\citep{autotool}, while daVinci assigns a dedicated selection agent its own advantage
\citep{davinci}. \method{} alters
neither the descriptions nor the architecture: it makes the single policy's own
mid-episode read a trained decision \emph{inside} the trajectory that policy already
generates, so selection is learned from the same rollouts that measure execution.

\paragraph{Where the training signal lands in a multi-turn trajectory.}
Outcome-rewarded policy gradients \citep{ppo,grpo} attach one advantage to every token
of a rollout, and a growing literature identifies that uniformity as the bottleneck for
agents \citep{casurvey}. Two results sharpen it: reward-correlated signal concentrates
on short action spans rather than the long reasoning around them \citep{actbottle}, and
the signal-to-noise of trajectory-level estimators decays as the fraction of
consequential turns falls \citep{drowning}. Remedies differ mainly in granularity ---
step-level process supervision \citep{prm}, per-turn credit
\citep{trace,tcpo,turnppo}, segmentation at natural tool-use boundaries
\citep{carl}, counterfactual per-token credit from sibling rollouts \citep{craft}, and
within-sequence token or segment reweighting \citep{strace,gear} --- while a parallel
line adds tool-specific signals: decoupled multi-objective GRPO \citep{toolomni},
information-gain step scores \citep{igsearch}, paired advantage channels for tool
contribution \citep{taco}, a conditional efficiency channel for tool invocation
\citep{hdpo}, branching and procedure-level credit at selected decision points
\citep{appo}, and empirically calibrated per-turn rewards \citep{rewardcal}. Agent
Lightning instead decomposes agent trajectories into transitions within a hierarchical
RL framework \citep{agentlightning}. Skill-oriented RL co-evolves reusable skills and
task policies \citep{skillrl,skillone,skillrise}, trains autonomous skill
internalisation from paired skill and no-skill rollouts \citep{skillc}, or learns
whether to invoke a skill from local preferences \citep{selskill}. These methods
demonstrate the value of finer credit, but they target different events and do not
isolate the identity tokens of a policy's mid-episode multiway skill read as a
disjoint, length-independent loss channel.
\method{} instead partitions that support into two disjoint channels: the read call is
removed from the task loss outright, its identity tokens carry an action-local utility
centred over the group's read actions, and equal per-batch loss mass renders the
decision's weight length-independent. Because that utility is not derived from the
outcome, it also sidesteps the impossibility result that binds outcome-reward weighting
schemes \citep{lengthimp}.

\section{Conclusion}
\label{sec:conclusion}

This paper introduces \method{}, which makes an agent's mid-episode choice of which
skill to read a trained decision by partitioning one trajectory's token support into
two disjoint credit channels: outcome credit reaching only the execution tokens, and
an action-local advantage reaching only the tokens that name the chosen skill. The
design follows from a measurement on a finished run's own training artifacts, where
that decision is shown to carry a vanishing and increasingly wrong-signed share of
the gradient while remaining worth eleven points of task success. Across five agentic
benchmarks \method{} lifts a 9B policy well past the same budget spent on outcome
reward alone and past reference models an order of magnitude larger, cutting exposure
to misleading candidates by two thirds while reading fewer skills. Its limits are
equally plain: each configuration is a single run, the method needs training tasks
whose correct skill is known, and an action-local scheme cannot credit an abstention.
More broadly, when one trajectory interleaves decisions of genuinely different kinds,
partitioning the token support is a cheap and verifiable alternative to redistributing
a single broadcast advantage.


\bibliographystyle{plainnat}
\bibliography{references}

\begin{thebibliography}{57}
\providecommand{\natexlab}[1]{#1}
\providecommand{\url}[1]{\texttt{#1}}
\expandafter\ifx\csname urlstyle\endcsname\relax
  \providecommand{\doi}[1]{doi: #1}\else
  \providecommand{\doi}{doi: \begingroup \urlstyle{rm}\Url}\fi

\bibitem[Anand and Chattaraj(2026)]{canary}
Atul Anand and Sourav Chattaraj.
\newblock Diagnosing tool-selection reasoning in llm agents with canary tools.
\newblock \emph{arXiv preprint arXiv:2608.04719}, 2026.

\bibitem[Chen et~al.(2026)Chen, Lin, Sun, Wang, Yang, Qin, Hu, Pan, and
  Zeng]{selskill}
Chishui Chen, Jiaye Lin, Te~Sun, Junxi Wang, Yi~Yang, Cong Qin, Yangen Hu,
  Lu~Pan, and Ke~Zeng.
\newblock Skill or skip? learning selective skill invocation in agentic tasks
  via dual-granularity preference learning.
\newblock \emph{arXiv preprint arXiv:2606.00510}, 2026.

\bibitem[Chen(2026)]{lookpick}
Shiyang Chen.
\newblock Looking is not picking: An attention-segment account of
  tool-selection failures in llm agents.
\newblock \emph{arXiv preprint arXiv:2606.16364}, 2026.

\bibitem[Cho et~al.(2026)Cho, Kang, and Kim]{skillret}
Hongcheol Cho, Ryangkyung Kang, and Youngeun Kim.
\newblock Skillret: A large-scale benchmark for skill retrieval in llm agents.
\newblock \emph{arXiv preprint arXiv:2605.05726}, 2026.

\bibitem[{DeepSeek-AI}(2025)]{deepseekr1}
{DeepSeek-AI}.
\newblock Deepseek-r1: Incentivizing reasoning capability in llms via
  reinforcement learning.
\newblock \emph{arXiv preprint arXiv:2501.12948}, 2025.

\bibitem[Ding et~al.(2026)Ding, Zhang, Liao, Zeng, and Yang]{lengthimp}
Fei Ding, Yongkang Zhang, Yuhao Liao, Zijian Zeng, and Huiming Yang.
\newblock On the impossibility of unbiased and length-invariant policy
  optimization with outcome rewards.
\newblock \emph{arXiv preprint arXiv:2607.23364}, 2026.

\bibitem[Ding(2026)]{skillresolvebench}
Jiandong Ding.
\newblock Skillresolve-bench: Measuring and resolving same-capability ambiguity
  in agent skill retrieval.
\newblock \emph{arXiv preprint arXiv:2606.10388}, 2026.

\bibitem[Esfandiarpoor et~al.(2025)Esfandiarpoor, Suryanarayanan, Bach,
  Chowdhary, and Aue]{mcpco}
Reza Esfandiarpoor, Vishwas Suryanarayanan, Stephen~H. Bach, Vishal Chowdhary,
  and Anthony Aue.
\newblock Themcpcompany: Creating general-purpose agents with task-specific
  tools.
\newblock \emph{arXiv preprint arXiv:2510.19286}, 2025.

\bibitem[Feng et~al.(2026)Feng, Wu, Gu, Lv, Jin, Zhang, Wen, and Tao]{taco}
Mingkuan Feng, Jinyang Wu, Hao Gu, Fangrui Lv, Ruihan Jin, Chuyuan Zhang,
  Zhengqi Wen, and Jianhua Tao.
\newblock Taco: Tool-augmented credit optimization for agentic tool use.
\newblock \emph{arXiv preprint arXiv:2606.30251}, 2026.

\bibitem[Fu et~al.(2026)Fu, Jiang, Wang, Yang, Hu, Liu, Hou, and Liu]{davinci}
Dayuan Fu, Mohan Jiang, Tongyu Wang, Dian Yang, Jiarui Hu, Liming Liu, Jinlong
  Hou, and Pengfei Liu.
\newblock davinci-kernel: Co-evolving skill selection, summarization, and
  utilization via rl for gpu kernel optimization.
\newblock \emph{arXiv preprint arXiv:2606.16497}, 2026.

\bibitem[Gao et~al.(2026)Gao, Li, Yuan, Ji, Ma, and Wang]{skillreduce}
Yudong Gao, Zongjie Li, Yuanyuan Yuan, Zimo Ji, Pingchuan Ma, and Shuai Wang.
\newblock Skillreducer: Optimizing llm agent skills for token efficiency.
\newblock \emph{arXiv preprint arXiv:2603.29919}, 2026.

\bibitem[Han et~al.(2026)Han, Xu, Liao, Wang, Jiang, Di, Lu, Hu, and
  Xiao]{skilluse}
Jinyi Han, Yuanjian Xu, Ying Liao, Xinyi Wang, Zishang Jiang, Zixiang Di,
  Fanyang Lu, Zhichao Hu, and Yanghua Xiao.
\newblock Skill-use: Can llms actually use skills in agentic harnesses?
\newblock \emph{arXiv preprint arXiv:2608.04828}, 2026.

\bibitem[He et~al.(2026{\natexlab{a}})He, Zhu, Zhou, Gu, Liu, Huang, Zou, Wipf,
  Yu, and Wu]{actbottle}
Langzhou He, Junyou Zhu, Yue Zhou, Zhengyao Gu, Junhua Liu, Wei-Chieh Huang,
  Henry~Peng Zou, David Wipf, Philip~S. Yu, and Qitian Wu.
\newblock Resolving action bottleneck: Agentic reinforcement learning informed
  by token-level energy.
\newblock \emph{arXiv preprint arXiv:2605.14558}, 2026{\natexlab{a}}.

\bibitem[He et~al.(2026{\natexlab{b}})He, Zhao, Wang, and Chen]{progdisc}
Yifeng He, Yinzhe Zhao, Jicheng Wang, and Hao Chen.
\newblock Is progressive disclosure all you need for long-context agents?
\newblock \emph{arXiv preprint arXiv:2607.17598}, 2026{\natexlab{b}}.

\bibitem[Huang et~al.(2026)Huang, Zhang, Hu, and Zhang]{toolomni}
Shouzheng Huang, Meishan Zhang, Baotian Hu, and Min Zhang.
\newblock Toolomni: Enabling open-world tool use via agentic learning with
  proactive retrieval and grounded execution.
\newblock \emph{arXiv preprint arXiv:2604.13787}, 2026.

\bibitem[Jimenez et~al.(2023)Jimenez, Yang, Wettig, Yao, Pei, Press, and
  Narasimhan]{swebench}
Carlos~E. Jimenez, John Yang, Alexander Wettig, Shunyu Yao, Kexin Pei, Ofir
  Press, and Karthik Narasimhan.
\newblock Swe-bench: Can language models resolve real-world github issues?
\newblock \emph{arXiv preprint arXiv:2310.06770}, 2023.

\bibitem[Kumar et~al.(2026)Kumar, Wu, and Suley]{carl}
Abhijit Kumar, Zoey Wu, and Mohit Suley.
\newblock Knowing when to ask: Segment-level credit assignment for llm tool
  use.
\newblock \emph{arXiv preprint arXiv:2605.27788}, 2026.

\bibitem[Lewis et~al.(2020)Lewis, Perez, Piktus, Petroni, Karpukhin, Goyal,
  K{\"u}ttler, Lewis, tau Yih, Rockt{\"a}schel, Riedel, and Kiela]{rag}
Patrick Lewis, Ethan Perez, Aleksandra Piktus, Fabio Petroni, Vladimir
  Karpukhin, Naman Goyal, Heinrich K{\"u}ttler, Mike Lewis, Wen tau Yih, Tim
  Rockt{\"a}schel, Sebastian Riedel, and Douwe Kiela.
\newblock Retrieval-augmented generation for knowledge-intensive nlp tasks.
\newblock \emph{arXiv preprint arXiv:2005.11401}, 2020.

\bibitem[Li et~al.(2025)Li, Zhou, Meng, Vadera, Li, and Li]{turnppo}
Junbo Li, Peng Zhou, Rui Meng, Meet~P. Vadera, Lihong Li, and Yang Li.
\newblock Turn-ppo: Turn-level advantage estimation with ppo for improved
  multi-turn rl in agentic llms.
\newblock \emph{arXiv preprint arXiv:2512.17008}, 2025.

\bibitem[Li et~al.(2026)Li, Huang, Liu, Li, Fu, Zhao, Bian, Zhang, Zhang, and
  Wang]{gear}
Sijia Li, Yuchen Huang, Zifan Liu, Yanping Li, Jingjing Fu, Li~Zhao, Jiang
  Bian, Ling Zhang, Jun Zhang, and Rui Wang.
\newblock Gear: Granularity-adaptive advantage reweighting for llm agents via
  self-distillation.
\newblock \emph{arXiv preprint arXiv:2605.11853}, 2026.

\bibitem[Liang et~al.(2026)Liang, Ma, Chen, Qian, Dai, Mao, Zhang, Lei, and
  Ou]{igsearch}
Zihan Liang, Yufei Ma, Ben Chen, Zhipeng Qian, Huangyu Dai, Lingtao Mao, Xuxin
  Zhang, Chenyi Lei, and Wenwu Ou.
\newblock Ig-search: Step-level information gain rewards for search-augmented
  reasoning.
\newblock \emph{arXiv preprint arXiv:2604.15148}, 2026.

\bibitem[Liao et~al.(2026)Liao, Chen, and Tang]{tcpo}
Sicong Liao, Zhi Chen, and Yaohua Tang.
\newblock Tcpo: Turn-level credit policy optimization.
\newblock \emph{arXiv preprint arXiv:2608.01667}, 2026.

\bibitem[Lightman et~al.(2023)Lightman, Kosaraju, Burda, Edwards, Baker, Lee,
  Leike, Schulman, Sutskever, and Cobbe]{prm}
Hunter Lightman, Vineet Kosaraju, Yura Burda, Harri Edwards, Bowen Baker, Teddy
  Lee, Jan Leike, John Schulman, Ilya Sutskever, and Karl Cobbe.
\newblock Let's verify step by step.
\newblock \emph{arXiv preprint arXiv:2305.20050}, 2023.

\bibitem[Lin et~al.(2026)Lin, Kuai, Xue, and Wang]{skillc}
Hongxiang Lin, Zhirui Kuai, Erpeng Xue, and Lei Wang.
\newblock Skillc: Learning autonomous skill internalization in llm agents via
  contrastive credit assignment.
\newblock \emph{arXiv preprint arXiv:2605.27899}, 2026.

\bibitem[Liu et~al.(2023)Liu, Yu, Zhang, Xu, Lei, Lai, Gu, Ding, Men, Yang,
  Zhang, Deng, Zeng, Du, Zhang, Shen, Zhang, Su, Sun, Huang, Dong, and
  Tang]{agentbench}
Xiao Liu, Hao Yu, Hanchen Zhang, Yifan Xu, Xuanyu Lei, Hanyu Lai, Yu~Gu,
  Hangliang Ding, Kaiwen Men, Kejuan Yang, Shudan Zhang, Xiang Deng, Aohan
  Zeng, Zhengxiao Du, Chenhui Zhang, Sheng Shen, Tianjun Zhang, Yu~Su, Huan
  Sun, Minlie Huang, Yuxiao Dong, and Jie Tang.
\newblock Agentbench: Evaluating llms as agents.
\newblock \emph{arXiv preprint arXiv:2308.03688}, 2023.

\bibitem[Luo et~al.(2025)Luo, Zhang, He, Wang, Zhao, Li, Qiu, and
  Yang]{agentlightning}
Xufang Luo, Yuge Zhang, Zhiyuan He, Zilong Wang, Siyun Zhao, Dongsheng Li,
  Luna~K. Qiu, and Yuqing Yang.
\newblock Agent lightning: Train any ai agents with reinforcement learning.
\newblock \emph{arXiv preprint arXiv:2508.03680}, 2025.

\bibitem[Meng and Chen(2026)]{craft}
Zibin Meng and Kani Chen.
\newblock Craft: Counterfactual credit assignment from free sibling rollouts
  for self-distilled agentic reinforcement learning.
\newblock \emph{arXiv preprint arXiv:2606.29476}, 2026.

\bibitem[Modecrua et~al.(2026)Modecrua, Kaewtawee, Pachtrachai, and
  Kraisingkorn]{rewardcal}
Wachiravit Modecrua, Krittanon Kaewtawee, Krittin Pachtrachai, and Touchapon
  Kraisingkorn.
\newblock Multi-turn reinforcement learning for tool-calling agents with
  iterative reward calibration.
\newblock \emph{arXiv preprint arXiv:2604.02869}, 2026.

\bibitem[Mou et~al.(2026)Mou, Zhuang, Chen, and Zhang]{strace}
Chaoli Mou, Zhan Zhuang, Xinning Chen, and Yu~Zhang.
\newblock Beyond uniform credit assignment: Selective eligibility traces for
  rlvr.
\newblock \emph{arXiv preprint arXiv:2605.05965}, 2026.

\bibitem[Patil et~al.(2023)Patil, Zhang, Wang, and Gonzalez]{gorilla}
Shishir~G. Patil, Tianjun Zhang, Xin Wang, and Joseph~E. Gonzalez.
\newblock Gorilla: Large language model connected with massive apis.
\newblock \emph{arXiv preprint arXiv:2305.15334}, 2023.

\bibitem[Pernot and Retault(2026)]{drowning}
Yann Pernot and Vi~Retault.
\newblock Drowning in routine: Signal dilution in multi-turn agent training.
\newblock \emph{arXiv preprint arXiv:2606.22164}, 2026.

\bibitem[Qin et~al.(2023)Qin, Liang, Ye, Zhu, Yan, Lu, Lin, Cong, Tang, Qian,
  Zhao, Hong, Tian, Xie, Zhou, Gerstein, Li, Liu, and Sun]{toolllm}
Yujia Qin, Shihao Liang, Yining Ye, Kunlun Zhu, Lan Yan, Yaxi Lu, Yankai Lin,
  Xin Cong, Xiangru Tang, Bill Qian, Sihan Zhao, Lauren Hong, Runchu Tian,
  Ruobing Xie, Jie Zhou, Mark Gerstein, Dahai Li, Zhiyuan Liu, and Maosong Sun.
\newblock Toolllm: Facilitating large language models to master 16000+
  real-world apis.
\newblock \emph{arXiv preprint arXiv:2307.16789}, 2023.

\bibitem[Rafailov et~al.(2023)Rafailov, Sharma, Mitchell, Ermon, Manning, and
  Finn]{dpo}
Rafael Rafailov, Archit Sharma, Eric Mitchell, Stefano Ermon, Christopher~D.
  Manning, and Chelsea Finn.
\newblock Direct preference optimization: Your language model is secretly a
  reward model.
\newblock \emph{arXiv preprint arXiv:2305.18290}, 2023.

\bibitem[Repantis et~al.(2026)Repantis, Gawde, Singh, and II]{slatesize}
Vyzantinos Repantis, Ameya Gawde, Harshvardhan Singh, and Joey~Blackwell II.
\newblock How many tools should an llm agent see? a chance-corrected answer.
\newblock \emph{arXiv preprint arXiv:2605.24660}, 2026.

\bibitem[Schick et~al.(2023)Schick, Dwivedi-Yu, Dess{\`i}, Raileanu, Lomeli,
  Zettlemoyer, Cancedda, and Scialom]{toolformer}
Timo Schick, Jane Dwivedi-Yu, Roberto Dess{\`i}, Roberta Raileanu, Maria
  Lomeli, Luke Zettlemoyer, Nicola Cancedda, and Thomas Scialom.
\newblock Toolformer: Language models can teach themselves to use tools.
\newblock \emph{arXiv preprint arXiv:2302.04761}, 2023.

\bibitem[Schulman et~al.(2017)Schulman, Wolski, Dhariwal, Radford, and
  Klimov]{ppo}
John Schulman, Filip Wolski, Prafulla Dhariwal, Alec Radford, and Oleg Klimov.
\newblock Proximal policy optimization algorithms.
\newblock \emph{arXiv preprint arXiv:1707.06347}, 2017.

\bibitem[Shao et~al.(2024)Shao, Wang, Zhu, Xu, Song, Bi, Zhang, Zhang, Li, Wu,
  and Guo]{grpo}
Zhihong Shao, Peiyi Wang, Qihao Zhu, Runxin Xu, Junxiao Song, Xiao Bi, Haowei
  Zhang, Mingchuan Zhang, Y.~K. Li, Y.~Wu, and Daya Guo.
\newblock Deepseekmath: Pushing the limits of mathematical reasoning in open
  language models.
\newblock \emph{arXiv preprint arXiv:2402.03300}, 2024.

\bibitem[Shi et~al.(2026)Shi, Chen, Lu, Miao, Liu, GU, Cai, Wang, and
  Zhang]{skillone}
Yaorui Shi, Yuxin Chen, Zhengxi Lu, Yuchun Miao, Shugui Liu, Qi~GU, Xunliang
  Cai, Xiang Wang, and An~Zhang.
\newblock Skill1: Unified evolution of skill-augmented agents via reinforcement
  learning.
\newblock \emph{arXiv preprint arXiv:2605.06130}, 2026.

\bibitem[Shinn et~al.(2023)Shinn, Cassano, Berman, Gopinath, Narasimhan, and
  Yao]{reflexion}
Noah Shinn, Federico Cassano, Edward Berman, Ashwin Gopinath, Karthik
  Narasimhan, and Shunyu Yao.
\newblock Reflexion: Language agents with verbal reinforcement learning.
\newblock \emph{arXiv preprint arXiv:2303.11366}, 2023.

\bibitem[Song and Wei(2026)]{shadowing}
Hongwen Song and Song Wei.
\newblock More skills, worse agents? skill shadowing degrades performance when
  expanding skill libraries.
\newblock \emph{arXiv preprint arXiv:2605.24050}, 2026.

\bibitem[Su et~al.(2026)Su, Long, Ai, He, Tang, Wang, Tu, Wang, and
  Liu]{skillra}
Weihang Su, Jianming Long, Qingyao Ai, Qiaozhi He, Yichen Tang, Changyue Wang,
  Yiteng Tu, Yingbo Wang, and Yiqun Liu.
\newblock Skill retrieval augmentation for agentic ai.
\newblock \emph{arXiv preprint arXiv:2604.24594}, 2026.

\bibitem[Sun et~al.(2026)Sun, Liu, Yan, Wang, and Weng]{when2tool}
Chung-En Sun, Linbo Liu, Ge~Yan, Zimo Wang, and Tsui-Wei Weng.
\newblock Llm agents already know when to call tools -- even without reasoning.
\newblock \emph{arXiv preprint arXiv:2605.09252}, 2026.

\bibitem[Tao et~al.(2026)Tao, Peng, Yao, Ge, Cheng, Wang, Gao, and Li]{trace}
Leitian Tao, Baolin Peng, Wenlin Yao, Tao Ge, Hao Cheng, Mike~Hang Wang,
  Jianfeng Gao, and Sharon Li.
\newblock Trace: Turn-level reward assignment via credit estimation for
  long-horizon agents.
\newblock \emph{arXiv preprint arXiv:2607.13988}, 2026.

\bibitem[Wang et~al.(2023)Wang, Xie, Jiang, Mandlekar, Xiao, Zhu, Fan, and
  Anandkumar]{voyager}
Guanzhi Wang, Yuqi Xie, Yunfan Jiang, Ajay Mandlekar, Chaowei Xiao, Yuke Zhu,
  Linxi Fan, and Anima Anandkumar.
\newblock Voyager: An open-ended embodied agent with large language models.
\newblock \emph{arXiv preprint arXiv:2305.16291}, 2023.

\bibitem[Wang et~al.(2026{\natexlab{a}})Wang, Ma, Wang, Ji, Yang, Chen, Wang,
  and Chu]{appo}
Xucong Wang, Ziyu Ma, Yong Wang, Yuxiang Ji, Shidong Yang, Guanhua Chen,
  Pengkun Wang, and Xiangxiang Chu.
\newblock Appo: Agentic procedural policy optimization.
\newblock \emph{arXiv preprint arXiv:2606.12384}, 2026{\natexlab{a}}.

\bibitem[Wang et~al.(2026{\natexlab{b}})Wang, Wen, Ji, and Qiao]{cappages}
Zifei Wang, Wei Wen, Qiang Ji, and Ruizhi Qiao.
\newblock Skills know their neighbors: Cluster-contrastive capability pages for
  skill retrieval.
\newblock \emph{arXiv preprint arXiv:2608.04482}, 2026{\natexlab{b}}.

\bibitem[Xia et~al.(2026)Xia, Chen, Wang, Liu, Zeng, Wang, Han, Zhou, Zhao,
  Chen, Zheng, Xie, and Yao]{skillrl}
Peng Xia, Jianwen Chen, Hanyang Wang, Jiaqi Liu, Kaide Zeng, Yu~Wang, Siwei
  Han, Yiyang Zhou, Xujiang Zhao, Haifeng Chen, Zeyu Zheng, Cihang Xie, and
  Huaxiu Yao.
\newblock Skillrl: Evolving agents via recursive skill-augmented reinforcement
  learning.
\newblock \emph{arXiv preprint arXiv:2602.08234}, 2026.

\bibitem[Xiao et~al.(2026)Xiao, Li, Li, Li, Jie, Liu, Jun, Wang, Tashi, and
  Yu]{skillsight}
Jinying Xiao, Bin Li, Xiaopeng Li, Jianling Li, Jiacheng Jie, Xiaodong Liu,
  Ma~Jun, Chao Wang, Nyima Tashi, and Jie Yu.
\newblock Skillsight: Calibrating generic content bias for skill retrieval.
\newblock \emph{arXiv preprint arXiv:2607.18785}, 2026.

\bibitem[Xie et~al.(2024)Xie, Zhang, Chen, Li, Zhao, Cao, Hua, Cheng, Shin,
  Lei, Liu, Xu, Zhou, Savarese, Xiong, Zhong, and Yu]{osworld}
Tianbao Xie, Danyang Zhang, Jixuan Chen, Xiaochuan Li, Siheng Zhao, Ruisheng
  Cao, Toh~Jing Hua, Zhoujun Cheng, Dongchan Shin, Fangyu Lei, Yitao Liu,
  Yiheng Xu, Shuyan Zhou, Silvio Savarese, Caiming Xiong, Victor Zhong, and Tao
  Yu.
\newblock Osworld: Benchmarking multimodal agents for open-ended tasks in real
  computer environments.
\newblock \emph{arXiv preprint arXiv:2404.07972}, 2024.

\bibitem[Yan et~al.(2026)Yan, Tong, Xue, Tang, Wang, Shi, Zhang, Li, and
  Zou]{hdpo}
Shilin Yan, Jintao Tong, Hongwei Xue, Xiaojun Tang, Yangyang Wang, Kunyu Shi,
  Guannan Zhang, Ruixuan Li, and Yixiong Zou.
\newblock Act wisely: Cultivating meta-cognitive tool use in agentic multimodal
  models.
\newblock \emph{arXiv preprint arXiv:2604.08545}, 2026.

\bibitem[Yang et~al.(2024)Yang, Jimenez, Wettig, Lieret, Yao, Narasimhan, and
  Press]{sweagent}
John Yang, Carlos~E. Jimenez, Alexander Wettig, Kilian Lieret, Shunyu Yao,
  Karthik Narasimhan, and Ofir Press.
\newblock Swe-agent: Agent-computer interfaces enable automated software
  engineering.
\newblock \emph{arXiv preprint arXiv:2405.15793}, 2024.

\bibitem[Yao et~al.(2022)Yao, Zhao, Yu, Du, Shafran, Narasimhan, and
  Cao]{react}
Shunyu Yao, Jeffrey Zhao, Dian Yu, Nan Du, Izhak Shafran, Karthik Narasimhan,
  and Yuan Cao.
\newblock React: Synergizing reasoning and acting in language models.
\newblock \emph{arXiv preprint arXiv:2210.03629}, 2022.

\bibitem[Yao et~al.(2026)Yao, Chen, Lu, Xu, Sun, Guo, Lu, Cai, Zhang, Han,
  Wang, Ye, Gu, Cai, Liu, and Shen]{skillrise}
Zhiyuan Yao, Yuxin Chen, Zhengxi Lu, Zishan Xu, Yueqing Sun, Yifu Guo, Yuquan
  Lu, Zhengzhou Cai, Kangning Zhang, Zhuowen Han, Zi-Han Wang, Ziang Ye, Qi~Gu,
  Xunliang Cai, Weiwen Liu, and Yongliang Shen.
\newblock Skillrise: Agentic reinforcement learning for cross-task skill
  evolution.
\newblock \emph{arXiv preprint arXiv:2607.26784}, 2026.

\bibitem[Zhang(2026)]{casurvey}
Chenchen Zhang.
\newblock From reasoning to agentic: Credit assignment in reinforcement
  learning for large language models.
\newblock \emph{arXiv preprint arXiv:2604.09459}, 2026.

\bibitem[Zheng et~al.(2026)Zheng, Zhang, Ma, Yu, Zhu, Wu, Xu, Dong, Zhu, Huang,
  and Yu]{skillrouter}
YanZhao Zheng, ZhenTao Zhang, Chao Ma, YuanQiang Yu, JiHuai Zhu, Yong Wu,
  Tianze Xu, Baohua Dong, Hangcheng Zhu, Ruohui Huang, and Gang Yu.
\newblock Skillrouter: Skill routing for llm agents at scale.
\newblock \emph{arXiv preprint arXiv:2603.22455}, 2026.

\bibitem[Zhou et~al.(2023)Zhou, Xu, Zhu, Zhou, Lo, Sridhar, Cheng, Ou, Bisk,
  Fried, Alon, and Neubig]{webarena}
Shuyan Zhou, Frank~F. Xu, Hao Zhu, Xuhui Zhou, Robert Lo, Abishek Sridhar,
  Xianyi Cheng, Tianyue Ou, Yonatan Bisk, Daniel Fried, Uri Alon, and Graham
  Neubig.
\newblock Webarena: A realistic web environment for building autonomous agents.
\newblock \emph{arXiv preprint arXiv:2307.13854}, 2023.

\bibitem[Zou et~al.(2025)Zou, Yang, Qi, Chen, Ai, Shen, He, and Wang]{autotool}
Jiaru Zou, Ling Yang, Yunzhe Qi, Sirui Chen, Mengting Ai, Ke~Shen, Jingrui He,
  and Mengdi Wang.
\newblock Autotool: Dynamic tool selection and integration for agentic
  reasoning.
\newblock \emph{arXiv preprint arXiv:2512.13278}, 2025.

\end{thebibliography}

\beginappendix
\Cref{app:unified} defines the 385-trial protocol and gives the raw counts behind
\cref{tab:main}; \cref{app:frontier} describes the interface controls for the
frontier references; \cref{app:claw147} reports the held-out Claw-Eval subset;
\cref{app:behaviour} completes the read-behaviour table; and
\cref{app:bootstrap} gives uncertainty on the controlled comparison.

\section{The 385-trial protocol}
\label{app:unified}

\paragraph{Composition.}
The evaluation combines two task sets. The non-Claw portion contains 56 tasks
drawn from SkillsBench, SETA, SWE and Terminal-Bench~2.0. Each is evaluated four
times, contributing 224 trials. The Claw-Eval portion
contains one trial for each of 161 tasks. Fourteen come from the repeated
evaluation set and the remaining 147 form a disjoint held-out set. Together the
two portions yield the 385 trials reported in \cref{tab:main}. Because Claw-Eval
has one trial per task whereas the other benchmarks have four, the pooled metric
is trial-level success rather than task-level pass@$R$.

\paragraph{Evaluation controls.}
Both portions use the same 16-candidate mixed-slate construction, a fixed
task-specific slate order, the same OpenClaw-derived system prompt, disabled
hidden thinking, a 65{,}536-token context window and each benchmark's own
grader. The task list, slate contents and decoding seed are fixed across
methods. Trained models use the tool interface seen during training; the
frontier references use the compatible interface described in
\cref{app:frontier}.

\paragraph{Aggregation.}
The two task sets were evaluated in separate batches and concatenated using the
task membership above. The rule is fixed across methods: no trajectory is
rescored, selected or discarded according to model outcome. For Skill-free RL,
the original records for the 14 repeated Claw-Eval tasks were unavailable, so
we use a complete rerun of the same model under identical tasks, slates, prompt
and decoding settings. This recovery changes neither the evaluated task set nor
any denominator. Other rows were either concatenated directly or evaluated on
the complete protocol in one run.

\begin{table}[ht]
\centering
\caption{Raw successes for the non-frontier rows of \cref{tab:main}. Column
headers give the fixed number of trials for each benchmark.}
\label{tab:merge}
\footnotesize
\setlength{\tabcolsep}{5pt}
\begin{tabular}{lcccccc}
\toprule
Method & Claw (161) & SB (32) & SETA (120) & SWE (40) & TB2 (32) & All (385) \\
\midrule
Qwen3.5-9B           & 72 & 0 & 29 & \phantom{0}6 & \phantom{0}3 & 110 \\
Qwen3.5-27B          & 87 & 3 & 51 & 16 & 11 & 168 \\
\midrule
SFT (RL init)        & 82 & 2 & 48 & 18 & \phantom{0}7 & 157 \\
Selection BC         & 84 & 5 & 52 & 20 & 11 & 172 \\
SelSkill-DPO         & 85 & 0 & 57 & 24 & 12 & 178 \\
\midrule
Skill-free RL        & 90 & 3 & 57 & 17 & 10 & 177 \\
\baserl{}            & 92 & 1 & 60 & 18 & 10 & 181 \\
\skillonerow{}       & 92 & 3 & 46 & 21 & 10 & 172 \\
Task-mask only       & 88 & 3 & 58 & 18 & 10 & 177 \\
\ourrow \textbf{\method{}} & \textbf{97} & \textbf{5} & \textbf{65} &
  \textbf{26} & \textbf{12} & \textbf{205} \\
\bottomrule
\end{tabular}
\end{table}

\section{Protocol for the frontier reference rows}
\label{app:frontier}

The frontier block of \cref{tab:main} uses the same tasks, skill slates, prompt
content, turn budget and graders as the trained rows. These models are capability
references rather than controlled baselines, and two interface adaptations prevent
the harness from measuring formatting quirks instead of task performance.

\paragraph{Native function calling.}
Each frontier model uses its provider's native function-calling interface. Our
manual schema is retained for the models trained with it, but imposing that schema
on an unrelated API model can cause otherwise valid calls to be rejected by the
parser. The available tools, observations and success criteria remain unchanged.

\paragraph{Loop detection.}
The agent loop stops repeated tool calls to prevent degenerate trajectories. For
the frontier rows, the duplicate-call signature retains a longer portion of the
arguments so that legitimate calls with a shared prefix are not mistaken for a
loop. The stopping threshold and all other rollout limits are unchanged, and the
same rule is applied to every frontier model.

We include only models that successfully dispatch tool calls and sustain a
multi-turn interaction in this environment. Models with systematic interface
failures or immediate termination are omitted rather than interpreted as weak task
capability. The frontier block should therefore be read as an absolute reference,
not as an exhaustive leaderboard.

\section{Per-method results on the 147 held-out Claw tasks}
\label{app:claw147}

The 147 corrected Claw tasks provide a clean generalisation test: no Claw task
appears in training, and these tasks are disjoint from the 14 Claw tasks in the
repeated protocol. Each is evaluated once per model, so this subset supports
task-level pass@1 directly. We report it separately here as well as including it
in \cref{tab:main}.

\paragraph{Oracle-skill verification.}
The oracle skills were inherited without successful-trajectory provenance, so it
was not known whether following each skill would solve its task. Reporting a
selection metric against unverified targets would measure the wrong thing. We
therefore checked every body against its task, fixtures and grader, and repaired
incorrect instructions. Recurring defects included wrong endpoints or response
fields, incorrect fixture assumptions, mishandled dates, unsafe operation ordering
and omitted deliverables. Skill names and descriptions were left unchanged: the
verification changed what a skill teaches, not what it advertises, so the selection
input is identical. All reported trials completed and passed checks for grader
availability, isolated benchmark state and the intended model checkpoint.

\begin{table}[ht]
\centering
\caption{Results on the 147 held-out Claw tasks, with one trial per task and the
standard mixed slate. Grader mean is the benchmark's partial-credit score. Oracle
and misleading reads are overlapping events and can sum past 100\%.}
\label{tab:claw147}
\footnotesize
\setlength{\tabcolsep}{6pt}
\begin{tabular}{lcccc}
\toprule
Method & pass@1 (\%) & Grader mean & Read oracle (\%) & Read misleading (\%) \\
\midrule
Qwen3.5-9B         & 42.9 & 0.562 & \phantom{0}7.5 & \phantom{0}6.8 \\
Qwen3.5-27B        & 53.1 & 0.672 & 13.6 & \phantom{0}4.1 \\
\midrule
SFT (RL init)      & 51.7 & 0.647 & 42.9 & 42.2 \\
Selection BC       & 51.7 & 0.641 & 71.4 & 23.8 \\
SelSkill-DPO       & 53.7 & 0.675 & 59.2 & 35.4 \\
\midrule
Skill-free RL      & 56.5 & 0.696 & 38.1 & 51.0 \\
\baserl{}          & 57.8 & 0.697 & 49.7 & 53.7 \\
Task-mask only     & 55.8 & 0.697 & 55.1 & 63.3 \\
\ourrow \textbf{\method{}} & \textbf{61.2} & \textbf{0.729} &
  \textbf{79.6} & \textbf{15.0} \\
\midrule
\multicolumn{5}{l}{\emph{Credit-design ablations (\cref{tab:ablation})}}\\
Group-level regret & 54.4 & 0.695 & 36.1 & 48.3 \\
Trajectory bonus   & 61.2 & 0.732 & 44.2 & 48.3 \\
Action credit      & 65.3 & 0.741 & 77.6 & 20.4 \\
\bottomrule
\end{tabular}
\end{table}

Three patterns are visible. Among the principal methods, \method{} is the only one
that leads task performance while simultaneously improving both oracle and misleading
exposure. Task-mask only raises oracle reading but raises misleading reading as well:
removing harmful outcome credit can make the policy read more, but does not make it
selective. Action credit without the single-read constraint performs strongly on this
held-out subset, but at the cost of more misleading reads. Because this comparison is
one trial per task from single training runs, we treat that difference as
distribution-specific rather than evidence that the single-read constraint is free
in every setting.

\section{Full read-behaviour table}
\label{app:behaviour}

\Cref{tab:main} reports oracle and misleading exposure on the 385-trial protocol,
so behaviour and outcome share a denominator there. \Cref{tab:behaviour} gives the
full behaviour breakdown on the repeated-trial protocol used by
\cref{tab:ablation,tab:selector} and \cref{fig:cost}. Absolute read rates are higher
on this subset because its repeated tasks permit longer interactions, but the
ordering between methods is unchanged.

\begin{table}[ht]
\centering
\caption{Read behaviour on the 280-trial protocol. Oracle and misleading exposure
are multi-label events and can sum past 100\%; reads/trial counts distinct skill
names. The untrained rows rarely read, so their low misleading exposure reflects
abstention rather than selectivity.}
\label{tab:behaviour}
\footnotesize
\setlength{\tabcolsep}{5pt}
\begin{tabular}{lccccc}
\toprule
Method & Any read (\%) & Oracle (\%) & Misleading (\%) &
  $P(\text{oracle}\mid\text{read})$ & Reads/trial \\
\midrule
Qwen3.5-9B          & 15.0 & \phantom{0}5.7 & \phantom{0}8.9 & 38.1 & 0.19 \\
Qwen3.5-27B         & 15.7 & \phantom{0}5.7 & \phantom{0}8.9 & 36.4 & 0.18 \\
\midrule
SFT (RL init)       & 94.6 & 37.9 & 61.8 & 40.0 & 1.34 \\
Selection BC        & 98.2 & 71.4 & 31.4 & 72.7 & 1.05 \\
SelSkill-DPO        & 96.4 & 66.1 & 51.8 & 68.5 & 1.59 \\
\midrule
Skill-free RL       & 80.0 & 35.4 & 55.0 & 44.2 & 1.25 \\
\baserl{}           & 96.4 & 54.3 & 69.6 & 56.3 & 1.88 \\
Task-mask only      & 98.9 & 48.9 & 73.6 & 49.5 & 1.98 \\
\ourrow \textbf{\method{}} & 98.2 & \textbf{83.9} & \textbf{21.8} &
  \textbf{85.5} & 1.11 \\
\bottomrule
\end{tabular}
\end{table}

\section{Uncertainty on the controlled comparison}
\label{app:bootstrap}

Tasks, not trials, are the resampling unit: each task's repeats are averaged first,
then tasks are resampled with replacement 50{,}000 times using the same task indices
for both methods. On the repeated-trial protocol, pooled trial success gives
$+7.9$ pp for \method{} over \baserl{} with a 95\% interval of
$[+2.1, +14.3]$, which excludes zero. Pooled task pass@4 gives $+7.1$ pp with
$[-1.4, +15.7]$, which does not. We therefore report the trial-level improvement
and do not claim significance for pass@4. Per-benchmark intervals rest on 8 to 30
tasks and are correspondingly wide; even the largest point estimate has an interval
that crosses zero.

\begin{figure}[ht]
\centering
\includegraphics[width=\linewidth]{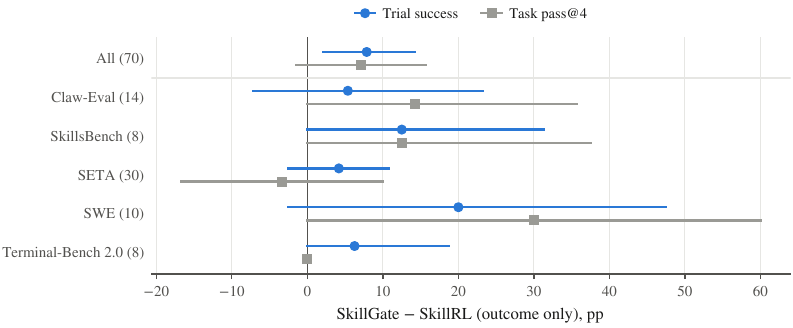}
\caption{\method{} minus \baserl{} per benchmark on the repeated-trial protocol.
Markers are observed differences and bars are 95\% task-clustered bootstrap
intervals. ``All'' is the pooled result, not a sixth benchmark.}
\label{fig:bootstrap}
\end{figure}

\section{Agent prompt and skill interface}
\label{app:prompt}

Training and evaluation use the same OpenClaw-derived prompt profile. It contains
the tool declarations, general agent instructions and a task-specific skill slate.
The operational sections unrelated to skill selection --- such as messaging,
memory and deployment administration --- are held fixed but are not reproduced
here. The released code contains the complete prompt builder. We instead show the
two parts that determine the selection action: the read-call surface and the
candidate representation.

\paragraph{Read-call surface.}
The trained models use a manual function-calling interface. A skill read has the
following form, where the path is copied from the selected slate entry:

\begin{lstlisting}[basicstyle=\ttfamily\scriptsize,breaklines=true,frame=single,
  columns=fullflexible,keepspaces=true,xleftmargin=2pt,xrightmargin=2pt]
<tool_call>
<function=read>
<parameter=path>
/skills/<skill-name>/SKILL.md
</parameter>
</function>
</tool_call>
\end{lstlisting}

The identity span in \cref{sec:setup} is the token span occupied by
\texttt{<skill-name>}. The surrounding function name and markup belong to the call
span but not to the identity span. This makes the selected candidate directly
observable in assistant-generated tokens.

\paragraph{Selection instruction.}
Before acting, the model sees the names, one-line descriptions and paths of all
16 candidates. The relevant prompt instruction is reproduced below in abridged
form; it is identical during training and evaluation.

\begin{lstlisting}[basicstyle=\ttfamily\scriptsize,breaklines=true,frame=single,
  columns=fullflexible,keepspaces=true,xleftmargin=2pt,xrightmargin=2pt]
Inspect the available skill descriptions before replying.
- If one skill clearly applies, read its SKILL.md and follow it.
- If several skills could apply, choose the most specific one.
- If none applies, do not read a skill.
Read no more than one skill up front, and use the exact listed path.
\end{lstlisting}

Each slate entry has the following structure:

\begin{lstlisting}[basicstyle=\ttfamily\scriptsize,breaklines=true,frame=single,
  columns=fullflexible,keepspaces=true,xleftmargin=2pt,xrightmargin=2pt]
<skill>
  <name>candidate-name</name>
  <description>one-line routing description</description>
  <location>/skills/candidate-name/SKILL.md</location>
</skill>
\end{lstlisting}

The slate order is fixed per task but hidden from the policy as metadata. One entry
is the oracle, five are misleading hard negatives and the remaining entries are
relevant or irrelevant bystanders, as defined in \cref{sec:exp:setup}. The model
sees no category label. Because hard negatives have similar names and descriptions,
the policy must resolve the choice from the task and these short descriptions before
opening any body. Requiring the exact listed path also makes off-slate and malformed
reads detectable, which is necessary for the token-level attribution used by
\method{}.

\end{document}